\documentclass[11pt,a4paper]{article}
\usepackage[hyperref]{emnlp2020}
\usepackage{times}
\usepackage{newtxtext,newtxmath}
\usepackage{latexsym}
\usepackage{mleftright}
\usepackage{multirow}
\usepackage{amsmath}
\usepackage{subcaption}
\usepackage{url}

\mleftright
\usepackage{amsfonts}

\usepackage{ctable}

\usepackage{microtype}

\aclfinalcopy 

\newcommand*{\affaddr}[1]{#1} 
\newcommand*{\affmark}[1][*]{\textsuperscript{#1}}

\title{Infusing Disease Knowledge into BERT for Health Question Answering,\\ Medical Inference and Disease Name Recognition}

\author{Yun He\affmark[1], Ziwei Zhu\affmark[1], Yin Zhang\affmark[1], Qin Chen\affmark[2], James Caverlee\affmark[1] \\
  \affaddr{\affmark[1]Texas A\&M University, College Station, USA} \\
  \affaddr{\affmark[2]Fudan University, Shanghai, China} \\
  \texttt{\{yunhe, zhuziwei, zhan13679, caverlee\}@tamu.edu}\\
  \texttt{qin\_chen@fudan.edu.cn}
  }

\date{}

\begin{document}
\maketitle
\begin{abstract}
Knowledge of a disease includes information of various aspects of the disease, such as signs and symptoms, diagnosis and treatment. This disease knowledge is critical for many health-related and biomedical tasks, including consumer health question answering, medical language inference and disease name recognition. While pre-trained language models like BERT have shown success in capturing syntactic, semantic, and world knowledge from text, we find they can be further complemented by specific information like knowledge of symptoms, diagnoses, treatments, and other disease aspects. Hence, we integrate BERT with disease knowledge for improving these important tasks. Specifically, we propose a new disease knowledge infusion training procedure and evaluate it on a suite of BERT models including BERT, BioBERT, SciBERT, ClinicalBERT, BlueBERT, and ALBERT. Experiments over the three tasks show that these models can be enhanced in nearly all cases, demonstrating the viability of disease knowledge infusion. For example, accuracy of BioBERT on consumer health question answering is improved from 68.29\% to 72.09\%, while new SOTA results are observed in two datasets. We make our data and code freely available.\footnote{\url{https://github.com/heyunh2015/diseaseBERT}}
\end{abstract}

\section{Introduction}

Human disease is ``a disorder of structure or function in a human that produces specific signs or symptoms'' \cite{diseaseDefinition}. Disease is one of the fundamental biological entities in biomedical research and consequently it is frequently searched for in the scientific literature \cite{islamaj2009understanding} and on the internet \cite{brownstein2009digital}.

\begin{table}[]
  \centering
\footnotesize
\setlength{\tabcolsep}{2.0pt}
\renewcommand\arraystretch{1.0}
  \caption{Disease knowledge of \textit{COVID-19} is presented from three aspects: symptoms, diagnosis and treatment (based on Wikipedia). }
\newcommand{\tabincell}[2]{\begin{tabular}{@{}#1@{}}#2\end{tabular}}
    \begin{tabular}{lll}
\toprule
    Disease & Aspect & Information \\
 \midrule
    \textit{COVID-19} & symptoms &  \tabincell{l}{Fever is the most common symptom, \\but highly variable in severity and \\presentation, with some older...} \\      
  \midrule
    \textit{COVID-19} & diagnosis &  \tabincell{l}{The standard method of testing is\\ real-time reverse transcription poly-\\merase chain reaction (rRT-PCR)...}\\
    \midrule
    \textit{COVID-19} & treatment &  \tabincell{l}{People are managed with supportive\\ care, which may include fluid therapy, \\oxygen support, and supporting...} \\
    \bottomrule
    \end{tabular}%
  \label{tab:Disease knowledge of COVID-19}%
\end{table}%

\begin{figure}[t]
\begin{subfigure}{.5\textwidth}
  \centering
 \setlength{\belowcaptionskip}{0.3cm}
  \includegraphics[width=1.0\linewidth]{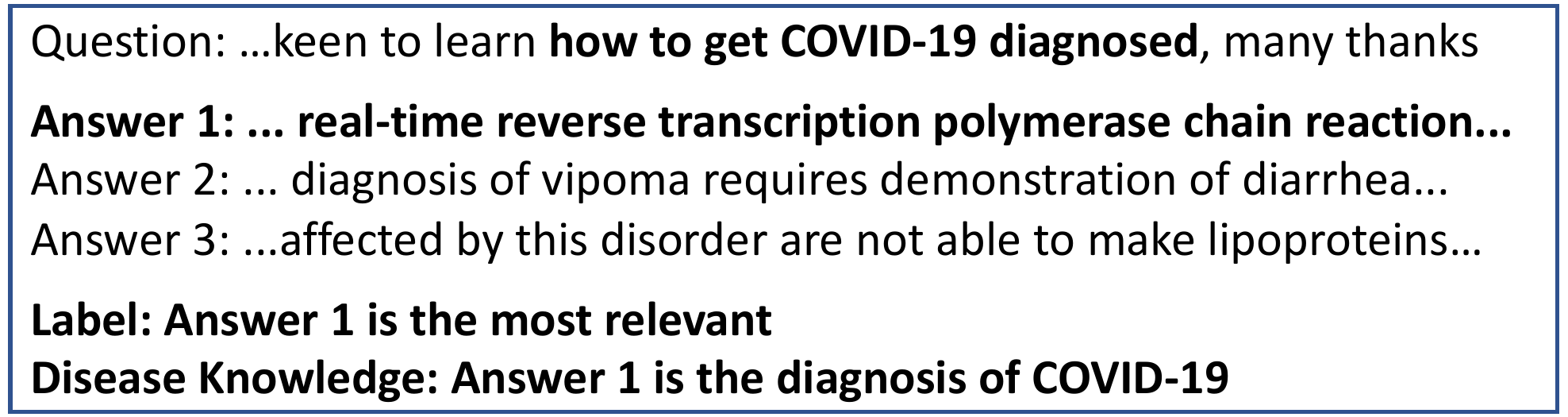}  
  \caption{Consumer Health Question Answering}
  \label{fig: examples of tasks - CHQ}
\end{subfigure}
\begin{subfigure}{.5\textwidth}
  \centering
 \setlength{\belowcaptionskip}{0.3cm}
  \includegraphics[width=1.0\linewidth]{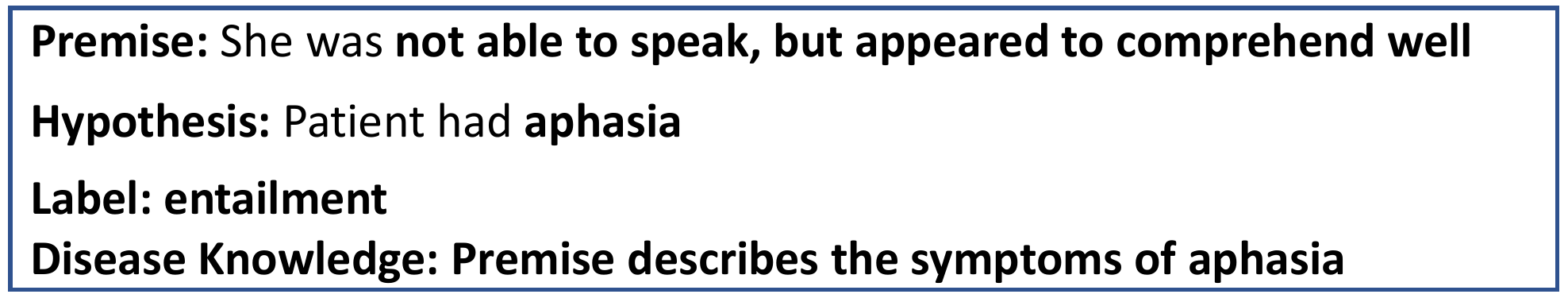}  
  \caption{Medical Language Inference}
  \label{fig: examples of tasks - MLI}
\end{subfigure}
\begin{subfigure}{.5\textwidth}
  \centering
  \includegraphics[width=1.0\linewidth]{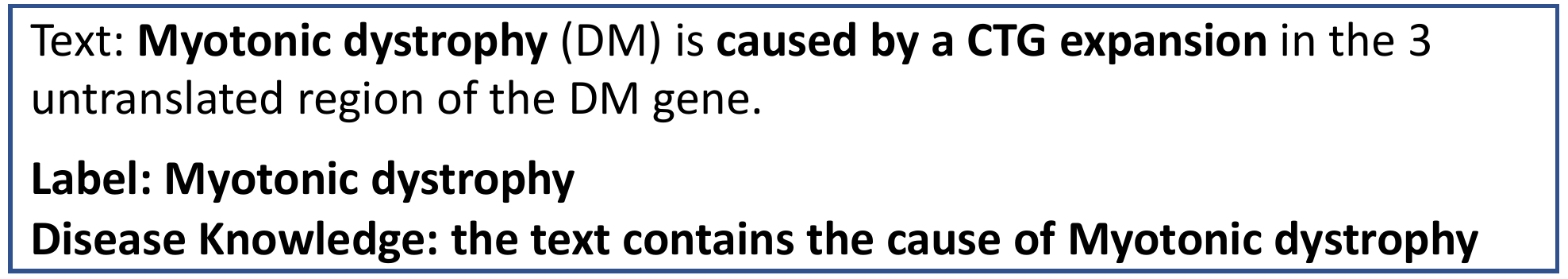}  
  \caption{Disease Name Recognition}
  \label{fig: examples of tasks - DNR}
\end{subfigure}
\caption{Examples of tasks that can benefit from disease knowledge.}
\label{fig: examples of tasks}
\end{figure}

Knowledge of a disease includes information about various aspects of the disease, like the signs and symptoms, diagnosis, and treatment \cite{saleem2012disease, urnes2008disease, du2017degree}. As an example, Table \ref{tab:Disease knowledge of COVID-19} highlights several aspects for COVID-19. Specialized disease knowledge is critical for many health-related and  biomedical natural language processing (NLP) tasks, including: \
\begin{itemize}
    \item \textit{Consumer health question answering} \cite{abacha2019overview} - the goal is to rank candidate passages for answering questions like ``What is the diagnosis of \textit{COVID-19}?'' as shown in Figure \ref{fig: examples of tasks - CHQ};
    \item \textit{Medical language inference} \cite{romanov-shivade-2018-lessons} - the goal is to predict if a given hypothesis (description of a patient) can be inferred from a given premise (another description of the patient);
    \item \textit{Disease name recognition} \cite{dougan2014ncbi} - the goal is to detect disease concepts in text.
\end{itemize}

For these tasks, it is critical for NLP models to capture disease knowledge, that is the semantic relations between a disease-descriptive text and its corresponding aspect and disease:
\begin{itemize}
    \item As shown in Figure \ref{fig: examples of tasks - CHQ}, if models can semantically relate ``...real-time reverse transcription polymerase chain reaction...'' (disease-descriptive text) to the diagnosis (aspect) of \textit{COVID-19} (disease), it is easier for them to pick up the most relevant answer among the candidates.
    \item Likewise, as shown in Figure \ref{fig: examples of tasks - MLI}, if models know that the premise is the symptoms (aspect) of \textit{Aphasia} (disease) in the hypothesis, they can easily predict that it is entailment not contradiction.
    \item Another example is shown in Figure \ref{fig: examples of tasks - DNR}, if models can semantically relate ``CTG expansion' to the cause (aspect) of \textit{Myotonic dystrophy} (disease), it is easier for them to detect this disease.
\end{itemize}
In a nutshell, NLP models require the disease knowledge for these disease-related tasks.

Recently, a new style of knowledge learning and leveraging has shaken NLP field with dramatic successes, enabled  by BERT \cite{devlin-etal-2019-bert} and its variants \cite{yang2019xlnet, liu2019roberta, raffel2019exploring, Lan2020ALBERT:}. These models capture language and world knowledge \cite{qiu2020pre, rogers2020primer} in their parameters via self-supervised pre-training over large-scale unannotated data and then leverage these knowledge in further fine-tuning over downstream tasks. Moreover, many biomedical BERT models such as BioBERT \cite{lee2020biobert} are proposed, which are pre-trained over biomedical corpora via a masked language model (MLM) that predicts randomly masked tokens given their context. This MLM strategy is designed to capture the semantic relations between random masked tokens and their context, but not the disease knowledge. Because the corresponding disease and aspect \textit{might not be randomly masked or might not be mentioned at all} in the disease-descriptive text, the semantic relations between them cannot be effectively captured via MLM. Therefore, a new training strategy is required to capture this disease knowledge.



In this paper, we propose a new \textit{disease knowledge infusion} training procedure to explicitly augment BERT-like models with the disease knowledge. The core idea is to train BERT to infer the corresponding disease and aspect from a disease-descriptive text, enabled by weakly-supervised signals from Wikipedia. Given a passage extracted from a section (normally describes an aspect) of a disease's Wikipedia article, BERT is trained to infer the title of the corresponding section (aspect name) and the title of the corresponding article (disease name). For example, in Table \ref{tab:Disease knowledge of COVID-19}, given ``...testing is real-time reverse transcription polymerase chain reaction (rRT-PCR)...'', BERT is trained to infer that this passage is from the section ``diagnosis" of the article ``COVID-19''. Moreover, because some passages do not mention the disease and aspect, we construct auxiliary sentences that contain the disease and aspect, such as ``What is the diagnosis of COVID-19?" and insert this sentence at the beginning of the corresponding passage. After that, we mask the disease and aspect in the auxiliary sentence and then let BERT-like models infer them given the passage. In this way, BERT learns how to semantically relate a disease-descriptive text with its corresponding aspect and disease.

To evaluate the quality of disease knowledge infusion, we conduct experiments on a suite of BERT models -- including BERT, BlueBERT, ClinicalBERT, SciBERT, BioBERT, and ALBERT -- over consumer health question (CHQ) answering, medical language inference, and disease name recognition. We find that (1) these models can be enhanced in nearly all cases. For example, accuracy of BioBERT on CHQ answering is improved from 68.29\% to 72.09\%; and (2) our method is superior to MLM for infusing the disease knowledge. Moreover, new SOTA results are observed in two datasets. These results demonstrate the potential of disease knowledge infusion into pre-trained language models like BERT.






\section{Related Work}

\noindent\textbf{Knowledge-Enriched BERT:} Incorporating external knowledge into BERT has been shown to be effective. Such external knowledge includes world (factual) knowledge for tasks such as entity typing and relation classification \cite{zhang2019ernie, peters2019knowledge, liu2019k, xiong2019pretrained}, sentiment knowledge for sentiment analysis \cite{tian2020skep, yin2020sentibert}, word sense knowledge for word sense disambiguation \cite{levine2019sensebert}, commonsense knowledge for commonsense reasoning \cite{klein2020contrastive} and sarcasm generation \cite{chakrabarty2020r}, legal knowledge for legal element extraction \cite{zhong2020does}, numerical skills for numerical reasoning \cite{geva2020injecting}, and coding knowledge for code generation \cite{xu2020incorporating}.

\noindent\textbf{Biomedical BERT:} BERT can also be enriched with biomedical knowledge via pre-training over biomedical corpora like PubMed, as in BioBERT \cite{lee2020biobert}, SciBERT \cite{beltagy2019scibert}, ClinicalBERT \cite{alsentzer2019publicly} and BlueBERT \cite{peng-etal-2019-transfer}. These biomedical BERT models report new SOTA performance on several biomedical tasks. Disease knowledge, of course, is a subset of biomedical knowledge. However, there are two key differences between these biomedical BERT models and our work: (1) Many biomedical BERT models are pre-trained via BERT's default MLM that predicts 15\% randomly masked tokens. In contrast, we propose a new training task: disease knowledge infusion, which infers the disease and aspect from the corresponding disease-descriptive text; (2) Biomedical BERT models capture the general syntactic and semantic knowledge of biomedical language, while our work is specifically designed for capturing the semantic relations between a disease-descriptive text and its corresponding aspect and disease. Experiments reported in Section~\ref{sec:experiments} show that our proposed method can improve the performance of each of these biomedical BERT models, demonstrating the importance of disease knowledge infusion.

\noindent\textbf{Biomedical Knowledge Integration Methods with UMLS:} Previous non-BERT methods connect data of downstream tasks with knowledge bases like UMLS \cite{sharma2019incorporating, romanov-shivade-2018-lessons}. For example, they map medical concepts and semantic relationships in the data to UMLS. After that, these concepts and relationships are encoded into embeddings and incorporated into models \cite{sharma2019incorporating}. The advantage is that they can explicitly incorporate knowledge into models. However, these methods have been outperformed by biomedical BERT models such as BioBERT in most cases.

%

\begin{table}[htbp]
  \centering
\footnotesize
\setlength{\tabcolsep}{2.7pt}
\renewcommand\arraystretch{1.0}
  \caption{Eight aspects of knowledge of a disease that are considered in this work.}
\newcommand{\tabincell}[2]{\begin{tabular}{@{}#1@{}}#2\end{tabular}}
    \begin{tabular}{ll}
\toprule
    Aspect Name & Definition \\
 \midrule
    Information & The general information of a disease. \\
    Causes & The causes of a disease. \\
    Symptoms & The signs and symptoms of a disease. \\
    Diagnosis & How to test and diagnose a disease. \\
    Treatment & How to treat and manage a disease. \\
    Prevention & How to prevent a disease. \\
    Pathophysiology & The physiological processes of a disease. \\
    Transmission & The means by which a disease spread. \\
    \bottomrule
    \end{tabular}%
  \label{tab:Eight aspects of a disease}%
\end{table}%

\section{Proposed Method: Disease Knowledge Infusion Training}
In this section, we propose a new training task: Disease Knowledge Infusion Training. Our goal is to integrate BERT-like pre-trained language models with disease knowledge to achieve better performance on a variety of medical domain tasks including answering health questions, medical language inference, and disease name recognition. Our approach is guided by three questions: Which diseases and aspects should we focus on? How do we infuse disease knowledge into BERT-like models? What is the objective function of this training task?

\subsection{Targeting Diseases and Aspects}
\label{Disease Terms }
First, we seek a disease vocabulary that provides disease terms. Several resources include Medical Subject Headings\footnote{\url{https://meshb.nlm.nih.gov/treeView}} (MeSH) \cite{lipscomb2000medical}, the National Cancer Institute thesaurus \cite{de2004nci}, SNOMED CT \cite{donnelly2006snomed}, and Unified Medical Language System (UMLS) \cite{bodenreider2004unified}. Each has a different scope and design purpose, and it is an open question into which is most appropriate here. As a first step, we select MeSH, which is a comprehensive controlled vocabulary proposed by the National Library of Medicine (NLM) to index journal articles and books in the life sciences, composed of 16 branches like anatomy, organisms, and diseases. We collect all unique disease terms from the Disease (MeSH tree number C01-C26) and Mental Disorder branch (MeSH tree number F01), resulting in 5,853 total disease terms.


\begin{figure*}[]
    \centering
    \includegraphics[scale=0.34]{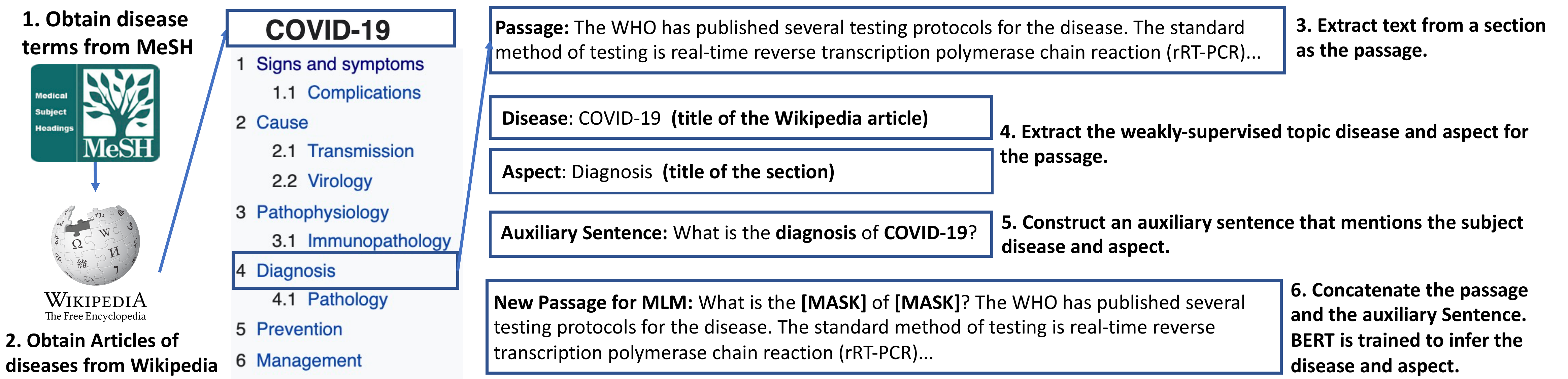}
    \caption{Disease Knowledge Infusion Training: An example with COVID-19.}
    \label{framework figure}
\end{figure*}

Knowledge of a disease involves information about various aspects of the disease \cite{saleem2012disease, urnes2008disease, du2017degree}. For each aspect, we focus on text alone (excluding images or other media). Following \citet{abacha2019question}, we consider eight disease aspects as shown in Table \ref{tab:Eight aspects of a disease}.

\subsection{Weakly Supervised Knowledge Infusion from Wikipedia}
\label{Weakly Supervision from Wikipedia}
Given the target set of diseases and aspects, the next challenge is how to infuse knowledge of the aspects of these diseases into BERT-like models. We propose to train BERT to infer the corresponding disease and aspect from a disease-descriptive text. By minimizing the loss between the predicted disease and aspect and the original disease and aspect, the model should memorize the semantic relations between the disease-descriptive text and its corresponding disease and aspect. 
 



A straightforward approach is to mask and predict the disease and aspect in the disease-descriptive text. However, this strategy faces two problems: (1) Given a passage extracted from disease-related papers, clinical notes, or biomedical websites, the ground-truth of its topic (i.e., disease and aspect) is difficult to identify. Medical expert annotation is time-consuming and expensive; while automatic annotation can suffer from large errors. For example, we need to recognize disease names in the passage, which is yet another challenging and still open problem in biomedical text mining  \cite{dougan2014ncbi}; (2) Diseases and aspects mentioned in a passage are not necessarily the topic words. Multiple disease names or aspect names might appear, making it difficult to determine which is the correct topic. For example, in Table \ref{tab:Disease knowledge of COVID-19}, the symptoms  of \textit{COVID-19} also mentions \textit{fever\footnote{Fever is included in the disease branch of MeSH.}}, while the correct topic is \textit{COVID-19}.



\medskip
\noindent\textbf{Weakly-Supervised Knowledge Source:} Instead of annotating an arbitrary disease-related passage, we exploit the structure of Wikipedia as a weakly-supervised signal. In many cases, each disease's Wikipedia article consists of several sections where each introduces an aspect of the disease (like diagnosis). For example, step 2 in Figure \ref{framework figure} shows several aspects on the Wikipedia page for \textit{COVID-19}. By extracting the passage from each section, the title of the section (e.g., diagnosis) is the topic aspect of the passage and the title of the article is the topic disease (e.g., COVID-19). Specifically, we search Wikipedia to obtain the articles for the 5,853 target disease terms from MeSH and apply regular expressions to extract the text of the sections corresponding to the appropriate aspects. In total, we collect a disease knowledge resource consisting of 14,617 passages.\footnote{Note that each disease article does not necessarily have all eight target aspects.} In fact, there are other online resources\footnote{\url{https://medlineplus.gov/skincancer.html}} with the similar structure. As a first step, we start with Wikipedia.


\medskip
\noindent\textbf{Auxiliary Sentences for Disease and Aspect Prediction:} The second problem is that the extracted passages do not necessarily mention the corresponding disease and the aspect. For example, in Table \ref{tab:Disease knowledge of COVID-19}, the disease name ``COVID-19'' does not appear in the information of its symptoms. In the disease knowledge resource, we find that only 51.4\% of passages mention both the corresponding diseases and aspects. Hence, we cannot simply mask-and-predict the disease and aspect because the passage does not mention them at all.

A remedy for this problem is an auxiliary sentence that contains the corresponding disease and aspect for each passage. We use a template of question style: ``What is the [\textit{Aspect}] of [\textit{Disease}]?'' to automatically generate auxiliary sentences as shown in step 5 in Figure \ref{framework figure}. Some examples are shown in Table \ref{tab:Examples of trigger sentences}. The advantage of this question style template is that the cloze statement of the auxiliary sentences for all aspects (except for the ``information'' aspect) are the same (What is the [MASK] of [MASK]?). Hence, the auxiliary sentences provide no clues (i.e., bias) for predicting the corresponding aspect.  

\begin{table}[htbp]
  \centering
\footnotesize
\setlength{\tabcolsep}{2.7pt}
\renewcommand\arraystretch{1.0}
  \caption{Examples of auxiliary sentences}
\newcommand{\tabincell}[2]{\begin{tabular}{@{}#1@{}}#2\end{tabular}}
    \begin{tabular}{ll}
\toprule
    Aspect Name & Auxiliary Sentence \\
\midrule
    Diagnosis & What is the diagnosis of COVID-19? \\
    Treatment & What is the treatment of COVID-19? \\
    Prevention & What is the prevention of COVID-19? \\
    Transmission & What is the transmission of COVID-19? \\
    Cloze Statement & What is the [MASK] of [MASK]? \\
    \bottomrule
    \end{tabular}%
  \label{tab:Examples of trigger sentences}%
\end{table}%


After that, we replace the corresponding disease and aspect with the special token [MASK] in the auxiliary sentences. Then, we insert the auxiliary sentence at the beginning of its corresponding passage to form a new passage with a question-and-answer style as shown in Figure \ref{framework figure}, where BERT is trained to predict the original tokens of the masked disease and aspect.


\subsection{Training Objective and Details}
Finally, we show the objective function of disease infusion training. Since most disease names are out of BERT vocabulary, the WordPiece tokenizer \cite{wu2016google} will split these terms into sub-word tokens that exist in the vocabulary. For example, ``COVID-19" will be split into 4 tokens: ``co", ``vid", ``-" and ``19". Formally, let $X = (x_{1}, ..., x_{T})$ denote a sequence of $T$ tokens that are split from a disease name where $x_{t}$ is the $t$-th token. The original cross-entropy loss is to get the conditional probability of a masked token as close as possible to the 1-hot vector of the token:

\begin{equation}
\setlength\abovedisplayskip{0pt}
	\mathcal{L}_{disease} = -\sum_{t=1}^{T}log  \ p(x_{t}|passage)
\end{equation}
where $p(x_{t}|context)$ is a conditional probability over $x_{t}$ given the corresponding passage, which can be defined as:

\begin{equation}
\setlength\abovedisplayskip{0pt}
\label{softmax equation}
	 p(x_{t}|passage) = \frac{exp(z_{t})}{\sum_{z \in \mathcal{V}} exp(z)}
\end{equation}
where $\mathcal{V}$ is the vocabulary and $z_{t}$ is the unnormalized log probability of $x_{t}$. Let $\textbf{y}_{t}$ denote the embedding of token $x_{t}$ from the output layer of BERT. We can estimate $z_{t}$ via:
\begin{equation}
\setlength\abovedisplayskip{1pt}
	z_{t} = \textbf{w}\cdot \textbf{y}_{t} + b
\end{equation}
where the weight $\textbf{w}$ and bias $b$ are learnable vectors. 

Note that the vocabulary size of BERT is around 30,000 which means masked language modeling task is a 30,000 multi-class problem. The logits (like $z_{t}$) after the normalization of softmax (Equation \ref{softmax equation}) will be pretty small (the expectation of mean should be around 1/30,000=3.3*e-5), which might cause some obstacles for the learning. Therefore, we also maximize the raw logits (like $z_{t}$) before softmax normalization which might keep more useful information. Empirically, we add the reciprocal of the logits to the cross-entropy loss:

\begin{equation}
\setlength\abovedisplayskip{0pt}
\label{final loss function}
	\mathcal{L}_{disease} = -\sum_{t=1}^{T}log  p(x_{t}|passage) + \frac{\beta}{\sum_{t=1}^{T}z_{t}}
\end{equation}
where $\beta$ balances the two parts of the loss. The final objective function is combined with the loss of the disease and aspect: $\mathcal{L} = \mathcal{L}_{disease} + \mathcal{L}_{aspect}$ where $\mathcal{L}_{aspect}=-log \ p(a|passage)$ and $a$ is the token of the aspect name. By minimizing this loss function, BERT can update its parameters to store the disease knowledge.


\section{Experiments}
\label{sec:experiments}

In this section, we examine disease knowledge infusion into six BERT variants over three disease-related tasks: health question answering, medical language inference, and disease name recognition. 

%
%

\medskip
\noindent\textbf{Reproducibility: } \textit{The code and data in this paper is released.\footnote{\url{https://github.com/heyunh2015/diseaseBERT}}} A model is firstly initialized with the pre-trained parameters from BERT or its variants and then is further trained by disease knowledge infusion to capture the disease knowledge. We use a widely used Pytorch implementation\footnote{\url{https://github.com/huggingface/transformers}} of BERT and Adam as the optimizer. We empirically set learning rate as 1e-5, batch size as 16 and $\beta$ as 10. Because MeSH (5,853 disease terms) is chosen as the disease vocabulary in our experiments, as a smaller vocabulary compared with others like UMLS (540,000 disease terms), we obtain a relatively small dataset of 14,617 passages. Hence, the training of disease knowledge infusion is as fast as fine-tuning BERT over downstream datasets, which takes 2-4 epochs to enhance BERT for a better performance on downstream tasks, which will be discussed in Section \ref{Learning Curve}. The training is performed on one single NVIDIA V100 GPU and it takes about 10 minutes to complete one training epoch using BERT-base architecture. The reproducibility for fine-tuning over downstream tasks will be detailed in Section \ref{sec: tasks}.


\subsection{BERT and its Biomedical Variants}
\label{sec: Pre-trained Language Models}
We consider six BERT models: two pre-trained over general language corpora (BERT and ALBERT) and four pre-trained over biomedical corpora (Clinical BERT, BioBERT, BlueBERT and SciBERT).

\medskip
\noindent\textbf{BERT} \cite{devlin-etal-2019-bert} is a multi-layer bidirectional Transformer encoder. Since the following biomedical versions of BERT are often based on the BERT-base architecture (12 layers and 768 hidden embedding size with 108M parameters), we choose BERT-base here for fair comparison. 


\noindent\textbf{ALBERT\footnote{\url{https://huggingface.co/albert-xxlarge-v2}}} \cite{Lan2020ALBERT:} compresses the architecture of BERT by factorized embedding parameterization and cross-layer parameter sharing. Via this compression, ALBERT can have a substantially higher capacity than BERT, with stronger performance on many tasks. We choose the maximum version ALBERT-xxlarge (12 layers and 4096 hidden embedding size with 235M parameters).

\noindent\textbf{BioBERT\footnote{\url{https://github.com/dmis-lab/biobert}}} \cite{lee2020biobert} is the first BERT pre-trained on biomedical corpora. It is initialized with BERT's pre-trained parameters (108M) and then further trained over PubMed abstracts (4.5B words) and PubMed Central full-text articles (13.5B words). We choose the best version BioBERT v1.1.

\noindent\textbf{ClinicalBERT\footnote{\url{https://huggingface.co/emilyalsentzer}}} \cite{alsentzer2019publicly} is a BERT model initialized from BioBERT v1.0 \cite{lee2020biobert} and further pre-trained over approximately 2 million notes in the MIMIC-III v1.4 database of patient notes \cite{johnson2016mimic}. We adopt the best performing version of ClinicalBERT (108M parameters) based on discharge summaries of clinical notes: Bio-Discharge Summary BERT.


\noindent\textbf{BlueBERT\footnote{\url{https://github.com/ncbi-nlp/bluebert}}} \cite{peng-etal-2019-transfer} is firstly initialized from BERT (108M parameters) and further pre-trained over a biomedical corpus of PubMed abstracts and clinical notes \cite{johnson2016mimic}.

\noindent\textbf{SciBERT\footnote{\url{https://huggingface.co/allenai/scibert\_scivocab\_uncased}}} \cite{beltagy2019scibert} is a BERT-base (108M parameters) model pre-trained on a random sample of the full text of 1.14M papers from Semantic Scholar \cite{ammar-etal-2018-construction}, with 18\% of papers from the computer science domain and 82\% from the biomedical domain. 


\begin{table}[t]
  \centering
	\footnotesize
	\setlength{\tabcolsep}{2.2pt}
\renewcommand\arraystretch{1.0}
  \caption{Summary of Tasks and Datasets. }
    \begin{tabular}{lccc}
	\toprule
    Datasets  & Train & Dev   & Test \\
	\midrule
     MEDIQA-2019 &  208 $(1,701)^{1}$ & 25 (234) & 150 (1,107) \\
     TRECQA-2017 &  254 (1,969) & 25 (234) & 104 (839) \\
	\midrule
    MEDNLI & $11,232^{2}$ & 1,395 & 1,422 \\
	\midrule
    BC5CDR-disease &  $4,182^{3}$ & 4,244 & 4,424 \\
    NCBI   & 5,145 & 787   & 960 \\
	\bottomrule
    \end{tabular}%
    \\
    {\raggedright 1, Questions with associated answers; 2, Pairs of premise and hypothesis; 3, Disease name mentions. \par}
  \label{tab:statistics of datasets}%
\end{table}%

\begin{table*}[htbp]
	\small
	\setlength{\tabcolsep}{2.5pt}
\renewcommand\arraystretch{1.0}
  \centering
  \caption{Experimental Results}
    \begin{tabular}{l|ccc|ccc|c|c|c}
\toprule
Tasks	& \multicolumn{6}{c|}{Consumer Health Question Answering} & \multicolumn{1}{c|}{NLI} & \multicolumn{2}{c}{NER} \\
\midrule
   Datasets       & \multicolumn{3}{c|}{MEDIQA-2019} & \multicolumn{3}{c|}{TRCEQA-2017} & MEDNLI & BC5CDR & NCBI \\
\midrule
   Metrics(\%)       & Accuracy & MRR   & Precision & Accuracy & MRR   & Precision & Accuracy & F1    & F1 \\
\midrule
    BERT  & 64.95 & 82.72 & 66.49 & 74.61 & 56.17 & 52.55 & 75.95 & 83.09 & 85.14 \\
    BERT + disease* & 66.40$\uparrow$ & 83.33$\uparrow$ & 68.94$\uparrow$ & 75.33$\uparrow$ & 56.41$\uparrow$ & 54.01$\uparrow$ & 77.29$\uparrow$ & 83.47$\uparrow$ & 86.81$\uparrow$ \\
\midrule
    BlueBERT & 65.13 & 81.50 & 67.35 & 74.26 & 48.40 & 52.55 & 82.21 & 85.73 & 87.78 \\
    BlueBERT + disease & 68.47$\uparrow$ & 81.17 & 71.57$\uparrow$ & 77.59$\uparrow$ & 50.96$\uparrow$ & 57.62$\uparrow$ & 83.90$\uparrow$ & 86.30$\uparrow$ & 87.79$\uparrow$ \\
\midrule
    ClinicalBERT & 67.30 & 84.78 & 70.59 & 77.00 & 52.56 & 56.62 & 81.50 & 84.90 & 87.25 \\
    ClinicalBERT + disease & 69.02$\uparrow$ & 88.94$\uparrow$ & 69.84 & 78.90$\uparrow$ & 54.97$\uparrow$ & 60.40$\uparrow$ & 81.65$\uparrow$ & 85.63$\uparrow$ & 87.22 \\
\midrule
    SciBERT & 68.47 & 84.47 & 68.07 & 77.23 & 54.57 & 57.54 & 80.94 & 86.16 & 87.24 \\
    SciBERT + disease & 73.35$\uparrow$ & 85.44$\uparrow$ & 76.28$\uparrow$ & 79.02$\uparrow$ & 56.57$\uparrow$ & 59.57$\uparrow$ & 82.14$\uparrow$ & 86.34$\uparrow$ & 88.30$\uparrow$ \\
\midrule
    BioBERT & 68.29 & 83.61 & 72.78 & 77.12 & 49.84 & 57.25 & 81.86 & 85.99 & 87.70 \\
    BioBERT + disease & 72.09$\uparrow$ & 87.78$\uparrow$ & 74.40$\uparrow$ & 78.43$\uparrow$ & 54.76$\uparrow$ & 58.45$\uparrow$ & 82.21$\uparrow$ & 86.52$\uparrow$ & 87.14 \\
\midrule
    ALBERT & 76.54 & 88.46 & 81.41 & 75.09 & \textbf{58.57} & 53.03 & 85.48 & 84.28 & 87.56 \\
    ALBERT + disease & \textbf{79.49}$\uparrow$ & 90.00$\uparrow$ & \textbf{84.02}$\uparrow$ & \textbf{80.10}$\uparrow$ & 57.21 & \textbf{62.40}$\uparrow$ & \textbf{86.15}$\uparrow$ & 84.71$\uparrow$ & 87.69$\uparrow$ \\
\midrule
    SOTA*  & 78.00 & \textbf{93.67} & 81.91 & 77.23 & 54.57 & 57.54 & 84.00 & \textbf{87.15} & \textbf{89.71} \\
\bottomrule
    \end{tabular}%
\\
{\raggedright * SOTA, state-of-the-art as of May 2020, to the best of our knowledge. \par}
{\raggedright * `` + disease" means that we train BERT via disease knowledge infusion training before fine-tuning. \par}
  \label{tab: experimental results}%
\end{table*}%

\subsection{Tasks}
\label{sec: tasks}
We test disease knowledge infusion over three biomedical NLP tasks. The dataset statistics are in Table \ref{tab:statistics of datasets}. For fine-tuning of BERT and its variants, the batch size is selected from [16, 32] and learning rate is selected from [1e-5, 2e-5, 3e-5, 4e-5, 5e-5].

\medskip
\noindent\textbf{Task 1: Consumer Health Question Answering.}
\label{section: Consumer Health Question Answering}
The objective of this task is to rank candidate answers for consumer health questions.

\noindent\textbf{Datasets.} We consider two datasets: MEDIQA-2019 \cite{ben-abacha-etal-2019-overview} and TRECQA-2017 \cite{abacha2017overview}.\footnote{\url{https://sites.google.com/view/mediqa2019}} MEDIQA-2019 is based on questions submitted to the consumer health QA system CHiQA\footnote{\url{https://chiqa.nlm.nih.gov/}}. TRECQA-2017 is based on questions submitted to the National Library of Medicine. Medical experts manually re-ranked the original retrieved answers and provide \textit{Reference \ Score} (1 to 11) and \textit{Reference \ Rank} (4: Excellent, 3: Correct but Incomplete, 2: Related, 1: Incorrect). 

\noindent\textbf{Fine-tuning.} MEDIQA-2019 and TRECQA-2017 are used as the fine-tuning dataset for each other. MEDIQA-2019 also contains a validation set for tuning hyper-parameters for both datasets. Following \citet{xu2019doubletransfer}, the task is cast as a regression problem where the target score is:
$
	score = \textrm{\textit{Reference \ Score}} - \frac{\textrm{\textit{Reference \ Rank}}-1}{m}
$
where $m$ is the number of candidate answers. Each question-answer pair is packed as a single sequence as the input for BERT. A single linear layer is on top of the output embedding of the special token [CLS] to generate the predicted score. MSE is adopted as the loss and we use Adam as the optimizer. All hyper-parameters are tuned on the validation set in terms of accuracy, where we set the batch size as 16 and learning rate as 1e-5.

\noindent\textbf{SOTA.} The state-of-the-art (SOTA) performance on MEDIQA-2019 is achieved by \citet{xu2019doubletransfer}, which is an ensemble method. Because TRECQA-2017 is fine-tuned on MEDIQA-2019, which is different from the original settings \cite{abacha2017overview} (BERT had not been proposed at that time), we use the best result of SciBERT among the BERT models as SOTA for TRECQA-2017. 

\medskip
\noindent\textbf{Task 2: Medical Language Inference.}
\label{section: Medical Language Inference}
The goal of this task is to predict whether a given hypothesis can be inferred from a given premise.

\noindent\textbf{Datasets.} MEDNLI \cite{romanov-shivade-2018-lessons} is a natural language inference dataset for the clinical domain.\footnote{\url{https://physionet.org/content/mednli/1.0.0/}} For each premise (a description of a patient) selected from clinical notes (MIMIC-III), clinicians generate three hypotheses: entailment (alternate true description of the patient), contradiction (false description of the patient), and neutral (alternate description that might be true).

\noindent\textbf{Fine-tuning.} Following \citet{peng-etal-2019-transfer}, we pack the premise and hypothesis together into a single sentence. A linear layer is on top of the output embedding of [CLS] to generate logits. Cross-entropy loss function is adopted, and we use Adam as the optimizer. All hyper-parameters are tuned on the validation set in terms of accuracy, where we set the batch size as 32 and learning rate as 1e-5.

\noindent\textbf{SOTA.} To the best of our knowledge, the state-of-the-art on MEDNLI is achieved by BlueBERT, reported in \citet{peng-etal-2019-transfer}.

\medskip
\noindent\textbf{Task 3: Disease Name Recognition.}
\label{section: Named Entity Recognition} This task is to detect disease names from free text.

\noindent\textbf{Datasets.} BC5CDR\footnote{\url{https://github.com/ncbi-nlp/BLUE\_Benchmark}} \cite{wei2016assessing} and NCBI\footnote{\url{https://www.ncbi.nlm.nih.gov/CBBresearch/Dogan/}} \cite{dougan2014ncbi} are collections of PubMed titles and abstracts. Medical experts annotate diseases mentioned in the collection. Since BC5CDR includes both chemicals and diseases, we focus on diseases in this dataset.

\noindent\textbf{Fine-tuning.} Following \citet{peng-etal-2019-transfer}, we cast this task as a token-level tagging (classification) problem, where each token is classified into three classes: B (beginning of a disease), I (inside of a disease) or O (out of a disease). Cross-entropy is adopted as the loss function and we use Adam as the optimizer. All hyper-parameters are tuned on the validation set in terms of F1, where we set the batch size as 32 and learning rate as 5e-5.

\noindent\textbf{SOTA.} The best performance is achieved by BioBERT v1.1, reported in \citet{lee2020biobert}\footnote{Although SciBERT reports a better result in NCBI, it uses a conditional random field on top of BERT, which is more complicated than the linear layer normally used in fine-tuning for BERT models including BioBERT.}. 

\subsection{Results}
The experimental results are presented in Table \ref{tab: experimental results}. We show each original model and its disease knowledge infused variant (e.g,. \textit{BERT} and \textit{BERT + disease}). We have two main findings: 


\medskip
\noindent\textbf{Effectiveness of Disease Infusion.} First, by infusing disease knowledge via our new training regimen, we see a significant improvement in nearly all cases. For example, \textit{ALBERT + disease} achieves 80.10\% in terms of accuracy which is superior to 75.09\% by ALBERT alone on TRECQA-2017. Standing on the shoulders of ALBERT, disease knowledge infusion leads to state-of-the-art results on MEDIQA-2019 and MEDNLI, to the best of our knowledge. Although BERT and ALBERT are pre-trained on all of  Wikipedia, including the articles of diseases, they might not pay enough attention to the disease part since Wikipedia is so large.  Hence, disease knowledge infusion that leverages the Wikipedia structure to capture the disease knowledge is a complement for BERT and ALBERT. Moreover, it is encouraging to see the improvements of disease knowledge infusion in biomedical BERT models, even though these variants are already pre-trained over large-scale biomedical corpora like PubMed with access to comprehensive disease information. This improvement demonstrates that the disease knowledge captured by our method -- that is, the semantic relations between a disease-descriptive text and its corresponding aspect and disease -- is different from the general linguistic knowledge in the biomedical domain captured by the randomly masked tokens prediction strategy of these biomedical BERT models. To sum up, the results show that the proposed disease knowledge infusion method can effectively complement BERT and its biomedical variants and hence improve the performance on health question answering, medical language inference,  and disease name recognition. 


\medskip
\noindent\textbf{Effectiveness of Biomedical BERT Models.} We also  observe that BERT models pre-trained on biomedical corpora outperform the same BERT architecture that is pre-trained on general language corpora. For example, BioBERT achieves 68.29\% in terms of accuracy on MEDIQA-2019 while BERT only obtains 64.95\%. This demonstrates that with the same model architecture, pre-training on biomedical corpora can capture more biomedical language knowledge that improves BERT for downstream biomedical tasks.\footnote{Note that our results for the biomedical BERT models in Table \ref{tab: experimental results} are slightly different from the results reported in the original papers that normally only provide a search range for hyper-parameters and not the specific optimal ones.}

In addition, we find that a high-capacity model like ALBERT can achieve similar performance as biomedical BERT models on TRECQA-2017, BC5CDR and NCBI, and even better performance on MEDIQA-2019 and MEDNLI. This observation might motivate new biomedical pre-trained models based on larger models like ALBERT-xxlarge.


\subsection{Ablation Study}
\label{Ablation Study}
We present the results of an ablation study on MEDIQA-2019 in Table \ref{tab:Ablation Study on MEDIQA-2019}. Similar results are observed on other datasets but omitted here due to the space limitation. We first remove ``Auxiliary Sentence''. That is, we remove the auxiliary question: ``What is the [\textit{Aspect}] of [\textit{Disease}]?'' and let BERT to predict the corresponding disease and aspect in the original passage if they appear. We observe worse results in terms of accuracy and precision, which shows that the auxiliary sentence is an effective remedy for the problem that some passages do not mention their disease and aspects. We also remove aspect prediction or disease prediction in the auxiliary sentence; both lead to worse results but removing disease prediction leads to a much lower performance. This shows that it is more important for BERT to infer the disease than the aspect from the passage. We also pre-train BERT on the same corpus (the disease-related passages) as our method. Following \citet{devlin-etal-2019-bert}, we randomly mask 15\% tokens in each sentence and let BERT to predict them. As shown in ``15\% Randomly Masked Tokens", we observe that our proposed disease infusion training task outperforms the default masked language model in BERT. This shows that our approach that leverages the structure of Wikipedia article to enhance the disease knowledge infusion works better than simply adding more data to the training process. Specifically, via leveraging the Wikipedia structure, we could effectively mask key words like aspect names and disease names that are related to disease knowledge and hence more effective than randomly masking strategy over the simply added data.



\begin{table}[t]
  \centering
	\small
	\setlength{\tabcolsep}{2.0pt}
	\renewcommand\arraystretch{1.0}
  \caption{Ablation Study on MEDIQA-2019}
    \begin{tabular}{lccc}
	\toprule
    Variants & \multicolumn{1}{l}{Accuracy} & \multicolumn{1}{l}{MRR} & \multicolumn{1}{l}{Precision} \\
	\midrule
    Default & 79.49 & 90.00 & 84.02 \\
    - Auxiliary Sentence & 78.23 & 90.89 & 78.10 \\
    - Aspect Prediction & 78.41 & 89.06 & 80.00 \\
    - Disease Prediction & 72.90 & 85.72 & 79.44 \\
    15\% Randomly Masked Tokens & 77.06 & 87.33 & 85.18 \\
	\bottomrule
    \end{tabular}%
  \label{tab:Ablation Study on MEDIQA-2019}%
\end{table}%

\subsection{Learning Curve}
\label{Learning Curve}
\begin{figure}[]
    \centering
    \includegraphics[scale=0.16]{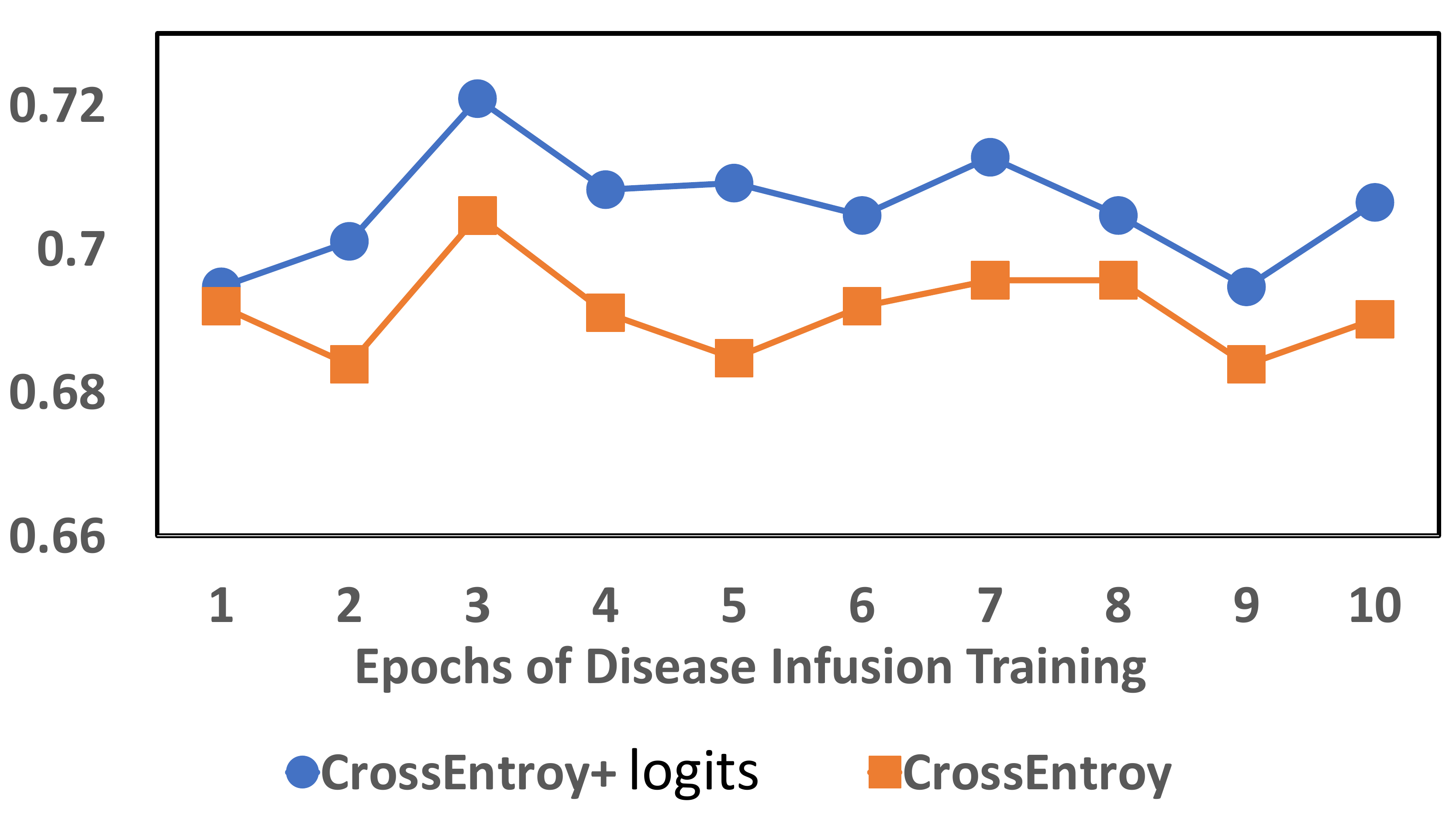}
    \caption{Learning curve of disease infusion knowledge. The y-axis is the accuracy of BERT models over MEDIQA-2019.}
    \label{fig: Learning curve of disease infusion knowledge}
\end{figure}

In this section, we present the learning curve of our proposed disease infusion training task. The x-axis denotes the  training epochs and the y-axis denotes the performance of BERT models that are augmented with disease infusion training at that epoch. We take BioBERT and MEDIQA-2019 as examples; similar results are obtained in other models over other tasks. The results in terms of accuracy are presented in Figure \ref{fig: Learning curve of disease infusion knowledge}, where we observe that (1) disease knowledge infusion takes only three epochs to achieve the optimal performance on BioBERT over the CHQ answering task. (2) cross-entropy loss used by disease knowledge infusion can be enhanced by adding the term of maximizing the raw logits (Equation \ref{final loss function}).

\section{Conclusions}
In this paper, we propose a new disease infusion training procedure to augment BERT-like pre-trained language models with disease knowledge. We conduct this training procedure on a suite of BERT models and evaluate them over disease-related tasks. Experimental results show that these models can be enhanced by this disease infusion method in nearly all cases.

\bibliography{emnlp2020}
\bibliographystyle{acl_natbib}

\end{document}